\documentclass{article}

\usepackage{microtype}
\usepackage{graphicx}
\usepackage{booktabs} 
\graphicspath{ {figures/} }
\usepackage{amsthm}
\usepackage{url}
\usepackage{subcaption}
\usepackage{graphicx}
\usepackage{amsmath}
\usepackage{amssymb}
\usepackage{xcolor}
\usepackage{thmtools,thm-restate}

\usepackage{hyperref}

\usepackage[accepted]{arxiv}
\icmltitlerunning{Augment your batch: better training with larger batches}
\newcommand{\e}{{\mathbb{E}}}
\newtheorem{theorem}{Theorem}

\begin{document}

\twocolumn[
\icmltitle{Augment your batch: better training with larger batches}



\icmlsetsymbol{equal}{*}

\begin{icmlauthorlist}
\icmlauthor{Elad Hoffer}{tech,habana}
\icmlauthor{Tal Ben-Nun}{eth}
\icmlauthor{Itay Hubara}{tech,habana}
\icmlauthor{Niv Giladi}{tech}
\icmlauthor{Torsten Hoefler}{eth}
\icmlauthor{Daniel Soudry}{tech}
\end{icmlauthorlist}

\icmlaffiliation{tech}{Department of Electrical Engineering, Technion, Haifa, 320003, Israel}
\icmlaffiliation{habana}{Habana-Labs, Caesarea, Israel}
\icmlaffiliation{eth}{Department of Computer Science, ETH Zurich}

\icmlcorrespondingauthor{Elad Hoffer}{elad.hoffer@gmail.com}

\icmlkeywords{Machine Learning, ICML}

\vskip 0.3in
]



\printAffiliationsAndNotice{\icmlEqualContribution} 


\begin{abstract}
Large-batch SGD is important for scaling training of deep neural networks.
However, without fine-tuning hyperparameter schedules, the generalization of the model may be hampered.
%
%
We propose to use batch augmentation: replicating instances of samples within the same batch with different data augmentations. 
Batch augmentation acts as a regularizer and an accelerator, increasing both generalization and performance scaling. 
We analyze the effect of batch augmentation on gradient variance, and show that it empirically improves convergence for a wide variety of deep neural networks and datasets. 
Our results show that batch augmentation reduces the number of necessary SGD updates to achieve the same accuracy as the state-of-the-art.
Overall, this simple yet effective method enables faster training and better generalization by allowing more computational resources to be used concurrently.

 

%

\end{abstract}

\section{Introduction}

Deep neural network training is a computationally-intensive problem, whose performance is inherently limited by the sequentiality of the Stochastic Gradient Descent (SGD) algorithm. In a common variant of the algorithm, a batch of samples is used at each step for gradient computation, accumulating the results to compute the descent direction. Batch computation enables \textit{data parallelism} \cite{bennun18demystifying}, which is necessary to scale training to a large number of processing elements.

Increasing batch size while mitigating accuracy degradation is actively researched in the ML and systems communities~\cite{goyal2017accurate,kurth17pf,jia18highly,mikami18224,osawa18kfac,ying18tpu}. \citet{shallue18dpar} comprehensively study the relation between batch size and convergence, whereas other works focus on increasing parallelism for a specific setting or hardware. Using such techniques, it is possible to reduce the time to successfully train ResNet-50~\cite{he2016deep} on the ImageNet~\cite{imagenet_cvpr09} dataset down to $132$ seconds~\cite{ying18tpu}, to the point where the performance bottleneck is reported to be input data processing (I/O) time.

The key to supporting large batch training often involves fine-tuning the base Learning Rate (LR), per-layer LRs~\cite{you2017scaling}, LR schedules (also called \textit{regimes})~\cite{goyal2017accurate,you2017scaling}, or the optimization step~\cite{krishnan2018neumann,hoffer2017train,osawa18kfac}. These methods typically use higher LRs to account for the lower gradient variance in large batch updates. However, without such fine-tuning, large batch training often results in degraded generalization. It was suggested this degradation is caused by a tendency of such low variance updates to converge to ``sharp minima'' \citep{keskar2016large}. 

 \begin{figure}[t]
 	\centering
 	\includegraphics[width=.46\textwidth,trim={0 0 0 0cm},clip]{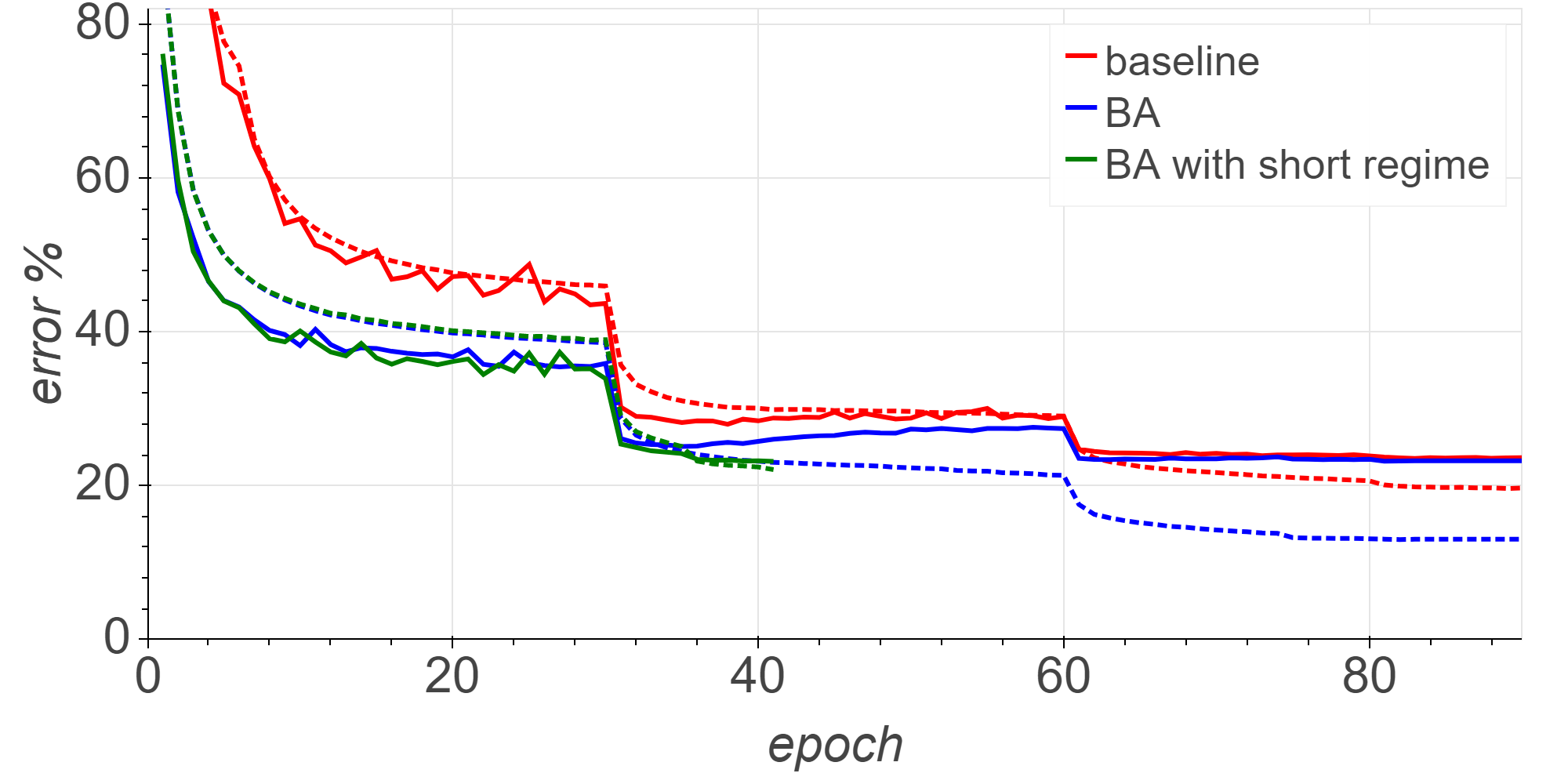}\vspace{-1em}
 	\caption{Impact of Batch Augmentation (BA, with M=4) on ResNet-50 and ImageNet, showing training (dashed) and validation error (solid).}
 	\label{compare_resnet50}
 	\label{compare_resnet50:baseline}\vspace{-1.5em}
 \end{figure} 
 
In this work, we propose \textbf{Batch Augmentation} (BA), which enables to control the gradient variance while increasing batch size. Using larger augmented batches, we can better utilize the computational resources~\cite{smith18dont,shallue18dpar} without the cost of additional I/O. In fact, it is even possible to achieve better generalization accuracy while adopting existing, standard LR schedules, i.e., without increasing learning rate, as can be seen in Figure~\ref{compare_resnet50}.

Our main contributions are:\\
\indent~~~~$\bullet$~~A formal analysis of batch augmentation.\\
\indent~~~~$\bullet$~~Empirical results for data augmentation properties,\\
\indent~~~~~~~~resource utilization, and gradient variance.\\
\indent~~~~$\bullet$~~Convergence results on multi-GPU nodes and a Cray\\
\indent~~~~~~~~supercomputer with 5,704 GPUs.

\subsection{Large batch training of neural networks}



%

Recent approaches by \citet{hoffer2017train}, \citet{goyal2017accurate}, \citet{you2017scaling} and others show that by adapting the optimization regime (i.e., hyperparameter schedule), large batch training can achieve equally good (and sometimes even better) generalization as training with small batches. 

\citet{hoffer2017train} argue that the quality of the optimized model stems from the number of SGD iterations, rather than the number of cycles through training data (epochs), and increase the number of steps w.r.t. the batch size. They then train ImageNet without accuracy degradation using additional epochs, adapting the points in which LR is reduced (Regime Adaptation), and normalizing subsets of the batch in a process called Ghost Batch Normalization (GBN).

\citet{goyal2017accurate} use batch size 8,192 and adopt a ``gradual warmup'' scheme, in which LR starts at zero and linearly increases to the base LR after 5 epochs, after which the regime resumes normally. \citet{you2017scaling} increases the batch size to 32,768 by using Layer-wise Adaptive Rate Scaling (LARS), as well as polynomial LR decay following warmup, with some reduction in accuracy. \citep{ying18tpu} employ distributed batch normalization (tuned for ghost batches of 64 images) and gradient accumulation to retain validation accuracy on ImageNet with 32,768 images per batch and 1,024 TPU devices. \citet{jia18highly} make use of 16-bit floating point (``half-precision'')  and further tune hyperparameters (e.g., weight decay) to reduce communication and enable training with batches of size 65,536.

Other large-batch methods include optimizers that utilize second-order information during training. The Neumann optimizer~\cite{krishnan2018neumann} uses a first-order approximation of the inverse Hessian using the Neumann Series, and is able to train up to batches of size 32,000 without accuracy degradation, albeit converging fastest when batches of 1,600 are used. The Kronecker Factorization (K-FAC) second-order approximation was also used to accelerate the convergence of deep neural network training~\cite{osawa18kfac}, achieving 74.9\% validation accuracy on ImageNet after 45 epochs, batch size of 32,768 on 1,024 nodes.





In contrast, \citet{masters2018revisiting} suggested that small batch updates may still provide benefits over large batch ones, showing better results over several tasks, with higher robustness to hyperparameter selection. The training process in this case, however, is sequential and cannot be distributed over multiple processing elements.

Batch Augmentation enables all benefits of large batch sizes while keeping the number of input examples constant and minimizing the number of hyperparameters. Furthermore, it improves generalization as well as hardware utilization. We now continue to discuss existing data augmentation techniques that we will later use for Batch Augmentation.


\subsection{A primer on data augmentation}
A common practice in training modern neural networks is to use data augmentation --- applying different transformations to each input sample. For example, in image classification tasks, for any input image, a random crop of varying size and scale is applied to it,  potentially together with rotation, mirroring and even color jittering  \citep{krizhevsky2012imagenet}. Data augmentations were repeatedly found to provide efficient and useful regularization, often accounting for significant portion of the final generalization performance \citep{Zagoruyko2016WRN, devries2017improved}. 

Several works even attempt to learn how to generate good data augmentations. For example,  Bayesian approaches based on the training set distribution \citep{tran2017bayesian}, generative approaches based on generative adversarial networks \citep{antoniou2017data,sixt2018rendergan} and search methods aimed to find the best data augmentation policy \citep{cubuk2018autoaugment}. Our approach is orthogonal to those methods, and we believe even better results can be obtained by combining them. 

Other regularization methods, such as Dropout \citep{srivastava2014dropout} or ZoneOut \citep{krueger2016zoneout}, although not explicitly considered as data augmentation techniques, can be considered as such by viewing them as random transforms over inputs for intermediate layers. These methods were also shown to benefit models in various tasks.
Another related regularization technique called "Mixup" was introduced by \citet{zhang2018mixup}. Mixup uses a mixed input from two separate samples with different classes, and uses as target their labels mixed by same amount. 

\section{Batch Augmentation} \label{sec:ba}
In this work, we suggest leveraging the merits of data augmentation together with large batch training, by using multiple instances of a sample in the same batch. 

We consider a model with a loss function $\ell(\mathbf{w},\mathbf{x}_n,\mathbf{y}_n)$ where $\left\{ \mathbf{x}_{n},\mathbf{y}_n\right\} _{n=1}^{N}$ is a dataset of
$N$ data sample-target pairs, where $x_n\in X$ and 
$T:X\rightarrow X$ is some data augmentation transformation applied to each example, e.g., a random crop of an image.
The common training procedure for each batch consists of the following update rule (here using vanilla SGD with a learning-rate $\eta$ and batch size of $B$, for simplicity):
\begin{align*}
    \mathbf{w}_{t+1}=\mathbf{w}_{t} -\eta \frac{1}{B}\sum_{n\in\mathcal{B}\left(k\left(t\right)\right)}  \nabla_\mathbf{w} \ell \left(  \mathbf{w}_t,T(\mathbf{x}_n), \mathbf{y}_n  \right)
\end{align*}
where $k\left(t\right)$ is sampled from $\left[N/B\right]\triangleq\left\{ 1,\dots,N/B\right\} $, $\mathcal{B}\left(k\right)$ is the set of samples in batch $k$, and we assume for simplicity that $B$ divides $N$.

We suggest to introduce $M$ multiple instances of the same input sample by applying the transformation $T_i$, here denoted by subscript $ i \in \left[M\right]$ to highlight the fact that they are different from one another.

We now use the slightly modified learning rule:
\begin{align*}
    \mathbf{w}_{t+1}=\mathbf{w}_{t} -\eta \frac{1}{M\cdot B}\sum_{i=1}^{M} \sum_{n\in\mathcal{B}\left(k\left(t\right)\right)} \nabla_\mathbf{w} \ell\left( \mathbf{w}_t, T_i(\mathbf{x}_n), \mathbf{y}_n \right)
\end{align*}
effectively using a larger $M\cdot B$ batch at each step, that is composed of $B$ samples augmented with $M$ different transforms each.

We note that this updated rule can be computed either by evaluating on the whole $M\cdot B$ batch or by accumulating $M$ instances of the original gradient computation. Using large batch updates as part of batch augmentations makes no change to the number of SGD iterations that are performed for each epoch. 

Batch augmentation (BA) can also be used to transform over intermediate layers, rather than just the inputs. For example, we can use the common Dropout regularization method \citep{srivastava2014dropout} to generate multiple instances of the same sample in a given layer, each with its Dropout mask. 

Batch augmentation can be easily implemented in any framework with reference PyTorch and TensorFlow implementations\footnote{To be available at \url{https://github.com/eladhoffer/convNet.pytorch}}.
To further highlight the ease of incorporating these ideas, we note that BA can be added to any training code by merely modifying the input pipeline -- augmenting each batch that is fed to the model.

\subsection{Hypothesis: large batch training issues}

In this section we examine the generalization issues with large-batch training. Then, in the next section, we provide insight into the increased generalization properties of BA.

Previous works \citep{keskar2016large,nar2018step,Wu2018sgd} suggested these issues may result from an implicit bias in the SGD training process: with large batch sizes, SGD selects minima with worse generalization. We examine the dyanmics of SGD to find how such a selection mechanism might work.

We examine the optimization of non-augmented datasets, using loss functions of the form 
\begin{equation}
f\left(\mathbf{w}\right)=\frac{1}{N}\sum_{n=1}^{N}\ell\left(\mathbf{w},\mathbf{x}_{n},\mathbf{y}_n\right)\,,\label{eq: f-1-1}
\end{equation}
where we recall $\left\{ \mathbf{x}_{n},\mathbf{y}_n\right\} _{n=1}^{N}$ is a dataset of
$N$ data sample-target pairs and $\ell$ is the loss function. We use SGD with batch of size $B$, where the update rule is given by
\begin{equation}
\mathbf{w}_{t+1}=\mathbf{w}_{t}-\eta \frac{1}{B}\sum_{n\in\mathcal{B}\left(k\left(t\right)\right)}\nabla_{\mathbf{w}}\ell \left(\mathbf{w}_t,\mathbf{x}_n,\mathbf{y}_n\right)\,.\label{eq: SGD-1}
\end{equation}
Here, we assume for simplicity that the indices are sampled with replacement,
$B$ divides $N$, and that $k\left(t\right)$ is sampled
uniformly from $\left\{ 1,\dots,N/B\right\} $. 
When our model is
sufficiently rich and over-parameterized (e.g., deep networks), we
typically converge to a minimum $\mathbf{w}^{*}$ which is a global
minimum on all data points in the training set~\citep{zhang2016understanding,soudry2017exponentially}. This means that $\forall n:\nabla_{\mathbf{w}}\ell\left(\mathbf{w}^{*},\mathbf{x}_n,\mathbf{y}_n\right)=0$.
We linearize the dynamics of Eq.~\ref{eq: SGD-1} near $\mathbf{w}^{*}$
to obtain
\begin{equation}
\mathbf{w}_{t+1}=\mathbf{w}_{t}-\eta\frac{1}{B}\sum_{n\in\mathcal{B}\left(k\left(t\right)\right)}\mathbf{H}_{n}\mathbf{w}_{t}\,\,,\label{eq: SGD-1-1}
\end{equation}
where we assume (without loss of generality) that $\mathbf{w}^{*}=0$,
and denote $\mathbf{H}_{n}\triangleq\nabla_{\mathbf{w}}^{2}\ell\left(\mathbf{w},\mathbf{x}_n,\mathbf{y}_n\right)$ as the per-sample Hessian.
Since we are at a global minimum, all $\mathbf{H}_{n}$ are symmetric PSD (there are no descent directions). However, recall that there can be many different global minima (on the training set). SGD selects only certain minima. As we shall see this selection depends on the batch sizes and learning rate, through the following quantities:
the averaged Hessian over batch $k$
	\[
	\left\langle \mathbf{H}\right\rangle _{k}\triangleq\frac{1}{B}\sum_{n\in\mathcal{B}\left(k\right)}\mathbf{H}_{n}
	\]
	and the maximum over the maximal eigenvalues of $\left\{ \left\langle \mathbf{H}\right\rangle _{k}\right\} _{k=1}^{N/B}$
	\begin{equation}
	\lambda_{\max}=\max_{k\in\left[N/B\right]}\max_{\forall\mathbf{v}:\left\Vert \mathbf{v}\right\Vert =1}\mathbf{v}^{\top}\left\langle \mathbf{H}\right\rangle _{k}\mathbf{v}\label{eq: lambda max-1}.
	\end{equation}

	This $\lambda_{\max}$ affects SGD through the following Theorem (proved in Appendix A):
	
\begin{theorem}
	\label{thm: maximal eigenvalue}
	The iterates of SGD (Eq.~\ref{eq: SGD-1-1})
	will converge if
	\begin{equation}
	\lambda_{\max}<\frac{2}{\eta}\,.\label{eq: stability 1-2}
	\end{equation}
	In addition, this bound is tight in the sense that it is also a necessary condition for certain datasets.
\end{theorem}

According to the Theorem, SGD with high learning rate will prefer to converge to minima with low $\lambda_{\max}$, thus selecting them from all (global) minima. Such minima, with low $\lambda_{\max}$, tend to have low variability of $\mathbf{H}_n$ (as high variability usually results in larger maximal values). 

Next, when increasing the batch size, we typically \textit{decrease} $\lambda_{\max}$, as we  \textit{decrease} the variability of $\left<\mathbf{H}\right>_k$ and replace max operations with averaging. Therefore, certain minima with high variability in $\mathbf{H}_n$ will suddenly become accessible to SGD. Now SGD may converge to these high variability minima, which were suggested to exhibit worse generalization performance than the original minima \citep{Wu2018sgd}. 

This issue can be partially mitigated by increasing the learning rate~\cite{hoffer2017train,goyal2017accurate}, in a way which will make these new minima inaccessible again, while keeping the original minima accessible. However, merely changing the learning rate may not be sufficient for very large batch sizes, when some minima with high variability and low variability will eventually have similar $\lambda_{\max}$, so SGD will not be able to discriminate between these minima. For example, in the limit of full batch (GD), the variability of $\mathbf{H}_n$ will not affect $\lambda_{\max}$ (only their mean).

\subsection{Data augmentation and variance reduction} 


Standard batch SGD averages the gradient over different samples, while BA additionally averages the gradient over several transformed instances $T\left(x_n\right)$ of the same samples. The augmented instances describe the same samples, typically with only small changes, and produce correlated gradients within the batch. 
As such, BA can achieve variance reduction that is significantly lower than the $1/\sqrt{B}$ reduction, which may occur with an uncorrelated sum of $B$ samples. 
This implies that the $\lambda_{\max}$ (Eq. \ref{eq: stability 1-2}) would change less in BA than standard large-batch training, allowing the model to 
exhibit less of the aforementioned SGD convergence issues.

In order to achieve this variance reduction, we must assume certain properties on $T$. Specifically, data augmentations should be designed such that they produce, in expectation, gradients that are more correlated with the original sample than other samples in the input dataset. More formally,
\begin{align*}
\e_{n\in \left[N\right]}\left[\mathrm{Corr}\left(\nabla_{\mathbf{w}}^{(n)},\nabla_{\mathbf{w}}\ell\left(\mathbf{w},T\left(x_n\right),y_n\right)\right)\right] \\
> \e_{n,m\in\left[N\right],\,n\neq m}\left[\mathrm{Corr}\left(\nabla_{\mathbf{w}}^{(n)},\nabla_{\mathbf{w}}^{(m)}\right)\right]\label{eq:aug}
\end{align*}
for $\nabla_{\mathbf{w}}^{(n)}\triangleq\nabla_{\mathbf{w}}\ell\left(\mathbf{w},x_n,y_n\right)$.
In the following section, we measure the effects of data augmentations used in practice and show that this property is maintained for standard image classification datasets.

\section{Characterizing Batch Augmentation}

We proceed to empirically study the different aspects of Batch Augmentation, including measurements of gradient correlation and variance, as well as performance and utilization analysis of augmented batches.

\subsection{Data augmentation characteristics}

To analyze the variance reduction of BA, we empirically show that data augmentations $T$ fulfill the assumption that they create correlated gradients in expectation. In particular, we measure the Pearson correlation coefficient $\rho$ between random images and augmented versions thereof $\rho\left(x,T\left(x\right)\right)$, as well as for random images of the same class $\rho\left(x,y\right)$  and different classes $\rho\left(z,w\right)$.

\begin{table}[ht]
    \vspace{-1em}
	\caption{ResNet-44 Gradient correlation on Cifar10. Augmentation types: \textbf{RC}=Random Crop, \textbf{F}=flip, \textbf{CO}=Cutout}
	\scriptsize
	\label{tab:correlation}
	\centering
	\begin{tabular}{lrrr}
		\toprule
		Measure & \multicolumn{3}{c}{Network State} \\
		\cmidrule{2-4}
		            & Init. & Partially & Fully \\
		            &       & Trained & Trained \\
		\midrule
		Epoch     & $0$ & $5$ & $93$ \\
        Val. Accuracy & $9.63\%$ & $63.24\%$ & $95.43\%$
        \\\addlinespace
        \midrule
        $\rho\left(x,T\left(x\right)\right)$ (RC,F)  & $0.99\pm 0$  & $0.56\pm 0.09$ & $0.13\pm 0.13$  \\
        $\rho\left(x,T\left(x\right)\right)$ (RC,F,CO) & $0.99\pm 0$  & $0.51\pm 0.08$ & $0.09\pm 0.08$ \\
        $\rho\left(x,y\right)$                &  $0.99\pm 0$  & $0.42\pm 0.06$ & $0.04\pm 0.03$ \\
        $\rho\left(z,w\right)$                &  -$0.11\pm 0.01$  & -$0.04\pm 0.06$ & $0 \pm 0.02$ \\
		\bottomrule
	\end{tabular}
	\vspace{-1em}
\end{table}

Table \ref{tab:correlation} lists the validation accuracy and median correlations (100 samples) between gradients of ResNet-44 on the Cifar10 dataset, at initialization, after $5$ epochs, and after convergence at $93$ epochs. In the table, it is clear to see that augmentations produce gradients that are considerably more correlated than images in different classes, and even within the same class. Moreover, the Cutout augmentation slightly decreases the gap between augmented and different images of the same class. As for the network state, when using random weights, interestingly all gradients of the same class correlate with each other. This suggests that at first, in expectation, there is a particular direction to descend to learn classifying a certain class of images, regardless of the actual sample. Correlation then decreases as training progresses. We now continue to measure the overall gradient variance when using augmented batches in training.

\subsection{Variance reduction characteristics}

As detailed in Section \ref{sec:ba}, using larger batches, both in standard practice and in BA, results in a smaller variance of the batch averaged version of the gradients and Hessians ($\left<\mathbf{H}\right>_k$). Therefore, in both cases $\lambda_{\max}$ decreases, in a way that may result in the large-batch issues described above --- the need to tune the learning rate, and the degraded performance with very large batch sizes. We now show that BA leads to significant variance reduction in practice. 

\begin{figure}[t]
	\centering
	\includegraphics[width=0.48\textwidth,trim={0 0 0 0},clip]{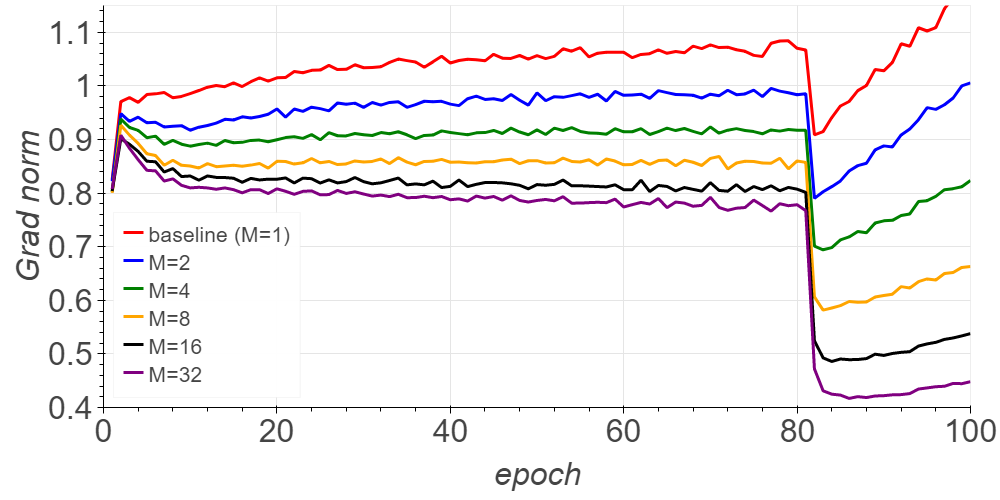}
	\caption{Comparison of gradient $L^2$ norm (ResNet44 + cutout, Cifar10, $B=64$) between the baseline ($M=1$) and batch augmentation with $M \in \{2,4,8,16,32\}$ }
	\label{grad_norm}
\end{figure}

In order to empirically evaluate this effect, we measured the $L^2$ of the gradients of the weights throughout the training for the setting described in Section~\ref{resnet44_m}. As could be expected, the variance reduction is reflected in the norm values as can be seen in Figure~\ref{grad_norm}. As the effective learning step is affected by this variance reduction, we can adapt the learning rate to partially account for this change, as described in the previous section. The correlated nature of the batch suggests that the needed learning rate correction for batch augmentation should be small.


\subsection{Performance characteristics}

The performance of BA is governed by two factors, which we study below: device utilization and work per iteration. 

In Table \ref{tab:tput}, we use one NVIDIA P100 GPU and the parallel filesystem of a Cray supercomputer to train the ImageNet dataset on ResNet-50 over all feasible batch sizes (limited by the device memory). We list the median values over 200 experiments of images processed per second, as well as standard deviation. As expected, increasing the batch size starts by scaling nearly linearly (1.8$\times$ between 1 and 2 images per batch), but slows scaling as we reach device capacity, with only 5.7\% utilization increase between batch sizes of 64 and 128. This indicates that, when using data parallelism in training, the local batch size should be increased as much as possible to maximize device utilization.

\begin{table}[ht]
    \vspace{-1em}
	\caption{ResNet-50 Image Throughput on ImageNet}
	\small
	\label{tab:tput}
	\centering
	\begin{tabular}{lrrr}
		\toprule
		Batch Size & Throughput & Standard\\
		           &[images/sec]& Deviation\\
		\midrule
        1  &     29.9 & 0.07 \\
        2  &     53.9 & 0.71 \\
        4  &     87.8 & 0.31 \\
        8  &    126.9 & 0.48 \\
        16 &    172.5 & 0.29 \\
        32 &    210.1 & 2.40 \\
        64 &    234.4 & 0.12 \\
        128&    247.9 & 0.12 \\
		\bottomrule
	\end{tabular}
	\vspace{-1em}
\end{table}

A theoretical understanding of the performance of parallel algorithms can be derived from the overall number of operations and the longest dependency path between them, which is a measure of the sequential part that fundamentally constrains the computation time (i.e., a work-depth model~\cite{blumofe99scheduling}). In BA and standard large-batch training, the overall number of operations (\textit{work}) increases proportionally to the overall batch size, i.e., $M\cdot B$. However, the sequential part (\textit{depth}), which is proportional to the number of SGD iterations, decreases as a result of faster LR schedules in BA, or shorter epochs in standard large-batch training. In essence, serialization can be reduced at the expense of more work, which increases the average parallelism.

Factoring for I/O and communication, BA also poses an advantage over standard large-batch training. BA decreases the dependency on external data, as in each iteration every processor can read the inputs and decode them once, applying augmentations locally. This increases scalability in state-of-the-art implementations, where input processing pipeline is the current bottleneck~\cite{ying18tpu}. Communication per iteration, on the other hand, is governed by the number of participating processing elements, in which the cost remains equivalent to standard large-batch training.

Our empirical results (e.g., Figure \ref{compare_resnet50}) show that in BA, the number of iterations may indeed be reduced as $M$ increases. This indicates that the time to completion can remain constant with better generalization properties. Thus, BA, in conjunction with large batches, opens an interesting tradeoff space between the work and depth of neural network training.

%
%

\section{Convergence analysis}
To evaluate the impact of Batch Augmentation (BA), we used several common datasets and neural network based models. 
For each one of the models, unless explicitly stated, we tested our approach using the original training regime and data augmentation described by its authors. To support our claim, we did not change the learning rate used nor the number of epochs.

\subsection{Cifar10/100}\label{resnet44_m}
The Cifar10 dataset introduced by \citet{krizhevsky2009learning} is a popular image classification dataset containing $50,000$ training images, together with a $10,000$ test set. Each image is of size $32 \times 32$ and belongs to one of $10$ classes of vehicles and animals. The Cifar100 dataset consists of the same number of training and validation images and the same spatial size, but with an increase to $100$ in the number of possible classes for each image.

For both datasets, we used the common data augmentation technique as described by \citet{he2016deep}. In this method, the input image is padded with $4$ zero-valued pixels at each side, top, and bottom. A random $32 \times 32$ part of the padded image is then cropped and with a $0.5$ probability flipped horizontally.  This augmentation method has a rather small space of possible transforms ($9\cdot 9 \cdot 2 = 162$), and so it is quickly exhausted by even a $M \approx 10$s of simultaneous instances.

We therefore speculated that using a more aggressive augmentation technique, with larger option space, will yield more noticeable difference when batch augmentation is used. We chose to use the recently introduced "Cutout" \citep{devries2017improved} augmentation method, that was noted to improve the generalization of models on various datasets considerably. Cutout uses randomly positioned zero-valued squares within images, thus increasing the number of possible transforms by $\times 30^2$.

We first tested batch augmentation on the task discussed by \citet{hoffer2017train} -- using a ResNet44 \citep{he2016deep} over the Cifar10 dataset \citep{krizhevsky2009learning} together with cutout augmentation \citep{devries2017improved}. We used the original regime by \citet{he2016deep} with a batch of $B=64$. We then compared the learning curve with training using batch augmentation with $M \in \{2,4,8,16,32\}$ different transforms for each sample in the batch, effectively creating a batch of $64 \cdot M$.

\begin{figure}[t!]
    \centering
    \begin{subfigure}[b]{0.48\textwidth}
        \includegraphics[width=\textwidth,trim={0 0 0 0},clip]{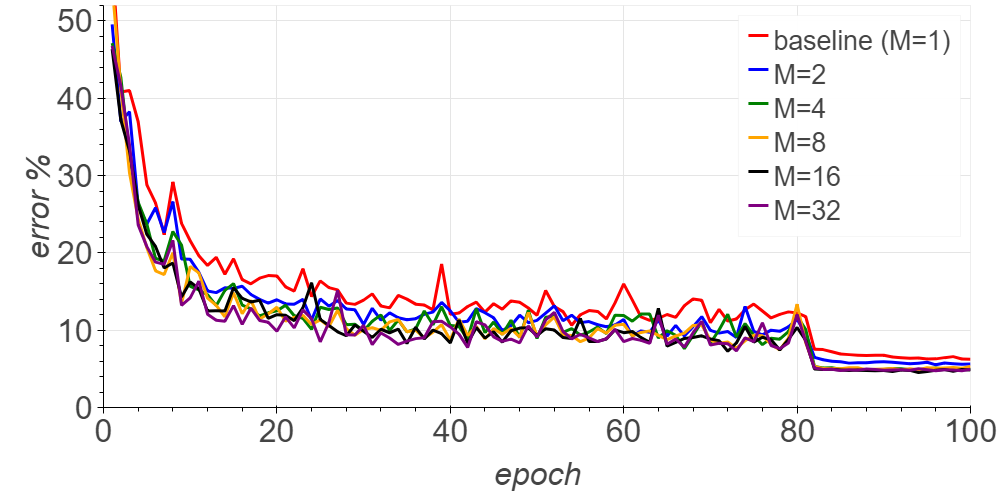}
        \caption{Validation error}
        \label{compare:baseline}
    \end{subfigure}
    ~ 
    \begin{subfigure}[b]{0.48\textwidth}
        \includegraphics[width=\textwidth,trim={0 0 0 0},clip]{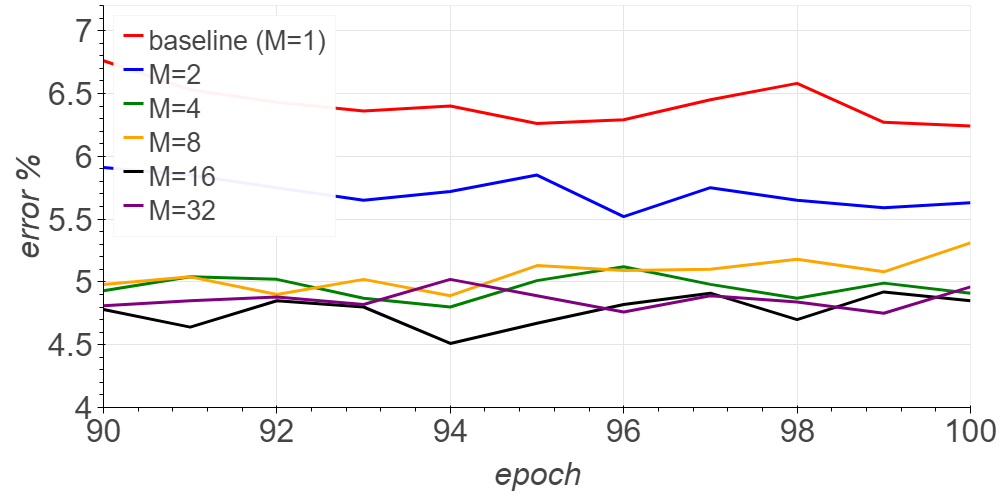}
        \caption{Final Validation error}
        \label{compare:zoom}
    \end{subfigure}\vspace{-1em}
    \caption{Impact of batch augmentation (ResNet44 + cutout, Cifar10). We used the original (red) training regime with $B=64$, and compared to batch augmentation with $M \in \{2,4,8,16,32\}$ creating an effective batch of $64\cdot M$}\label{compare}
\end{figure}

As we can see in Figure~\ref{compare}, validation convergence speed has noticeably improved (in terms of epochs), with a significant reduction in final validation classification error (Figure~\ref{compare:zoom}). This trend largely continues to improve as $M$ is increased, consistent with our expectation.


We verified these results using a variety of models (Table \ref{table:val_accuracy}) using various values of $M$, depending on our ability to fit the $M \cdot B$ within our compute budget. Our best result was achieved using DARTS final Cifar10 model \cite{liu2018darts}. DARTS is a differentiable architecture search framework that constructs a graph with SoftMax-parameterized edges. The final model is a subset of the graph whose edges have the highest values.

In all our experiments we have observed significant improvements to the final validation accuracy as well, as an increase in accuracy per epoch convergence speed. 



Moreover, we managed to achieve high validation accuracy much quicker with batch augmentation. We trained a ResNet44 with Cutout on Cifar10 for half of the iterations needed for the baseline, using batch augmentation, larger learning rate, and faster learning rate decay schedule. We managed to achieve $94.15\%$ accuracy in only $23$ epochs, whereas the baseline achieved $93.07\%$ with over four times the number of iterations ($100$ epochs). When the baseline is trained with the same shortened regime there is a significant accuracy degradation. This indicates not only an accuracy gain, but a potential runtime improvement for a given hardware.

We were additionally interested to verify that improvements gained with BA were not caused by simply viewing more sample instances during training. To make this distinction apparent, we compare with the training regime adaptation (RA) method by \citet{hoffer2017train}. In this method, the number of epochs is increased so that the number of iterations is fixed when using a larger batch. This makes both RA and BA methods comparable with respect to the number of instances seen for each sample over the course of training. Using the same settings (ResNet44, Cifar10), we find an accuracy gain of $0.5\%$ over the $93.07\%$ result reported by \citet{hoffer2017train}. Figure \ref{ba_ra} show these results. Additional results appear in supplementary material.

\begin{figure}[ht]
    \centering
    \begin{subfigure}[b]{0.48\textwidth}
        \includegraphics[width=\textwidth,trim={0 0 0 0cm},clip]{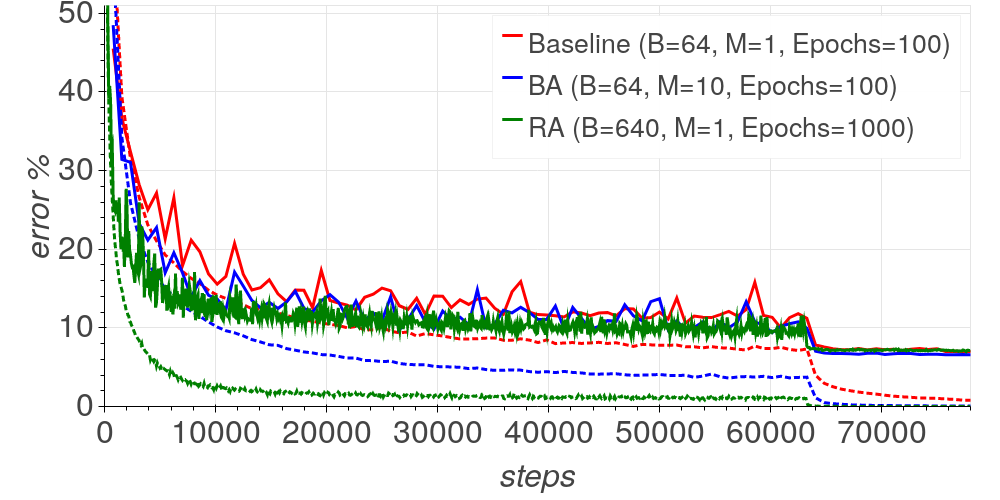}
        \caption{Training (dashed) and validation error}
        \label{ba_ra:baseline}
    \end{subfigure}
    ~ 
    \begin{subfigure}[b]{0.48\textwidth}
        \includegraphics[width=\textwidth,trim={0 0 0 0cm},clip]{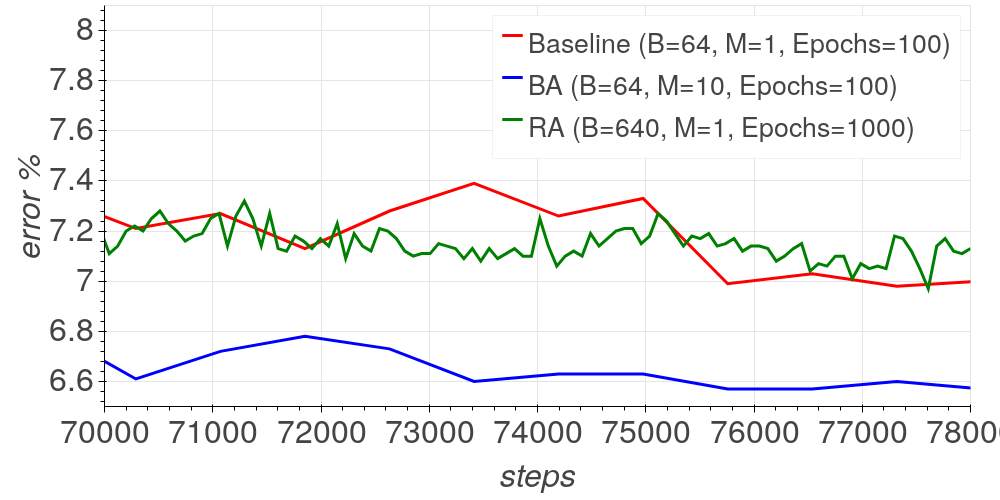}
        \caption{Training (dashed) and validation final error}
        \label{ba_ra:zoom}
    \end{subfigure}\vspace{-1em}
    \caption{ A comparison between 
(1) baseline B=64 training
(2) our batch augmentation (BA) method with M=10
(3) regime adaptation (RA) with B=640 and 10x more epochs.}\label{ba_ra}
  \end{figure}
  
\begin{table*}[ht]
\small
\centering{}
\caption{Validation accuracy (Top1) results for Cifar, ImageNet models. Bottom: test perplexity result on Penn-Tree-Bank (PTB) dataset. Relative change in error/ppl over baseline is listed in percentage, and improvements higher than 5\% are marked in bold.}
\label{table:val_accuracy}
\begin{tabular}{lrrrr}
\toprule{}\
Network                                       &   Dataset &   Baseline  & Batch Augmentation & Change  \\
\midrule
ResNet44 \citep{he2016deep}                   &   Cifar10 &   93.07\%    & 93.78\% (M=40) & \textbf{+10.24\%}\\
ResNet44 + cutout                             &   Cifar10 &   93.7\%     & 95.43\% (M=40) & \textbf{+27.46\%}\\
VGG + cutout \citep{simonyan2014very}         &   Cifar10 &   93.82\%    & 95.32\% (M=32) & \textbf{+24.27\%} \\
Wide-ResNet28-10 + cutout    \citep{Zagoruyko2016WRN} &  Cifar10  &   96.6\%     & 97.15\% (M=6) & \textbf{+16.18\%} \\
DARTS \citep{liu2018darts}                   &   Cifar10 &   97.11\%    & 97.64\% (M=10)  & \textbf{+18.34\%} \\
\midrule
ResNet44 + cutout                             &   Cifar100 &   72.97\%     & 74.13\% (M=40) & +4.2\%\\
VGG + cutout                                  &  Cifar100 &   73.03\%    & 75.5\% (M=32) & \textbf{+9.16\%} \\
Wide-ResNet28-10 + cutout     &  Cifar100 &   79.85\%     & 80.13\% (M=10) & +1.39\% \\
DenseNet100-12 \citep{huang2017densely}       &  Cifar100 &   77.73\%     & 78.8\% (M=4) & +4.8\%\\
\midrule
AlexNet \citep{krizhevsky2012imagenet}        &   ImageNet &   58.25\%     & 62.31\% (M=8) & \textbf{+9.72\%}\\
MobileNet \citep{howard2017mobilenets}        &   ImageNet &   70.6\%     & 71.4\% (M=4) & +2.72\%\\
ResNet50 \citep{he2016deep}                   &   ImageNet &   76.3\%     & 76.86\% (M=4) & +2.36\%\\
\midrule
Word-level LSTM \citep{merity2017regularizing}&   PTB &   58.8 ppl     & 58.6 ppl (M=10) & -0.3\%\\
\bottomrule
\end{tabular}
\end{table*}

\subsection{ImageNet}
As a larger scale evaluation, we used the ImageNet dataset \citep{imagenet_cvpr09}, containing more than 1.2 million images with 1,000 different categories.
We evaluate three models on the ImageNet task. For ResNet50 \citep{he2016deep}, we used the data augmentation method advocated by \citet{szegedy2015going} that employed various sized patches of the image with size distributed evenly between $8\%$ and $100\%$ and aspect ratio constrained to the interval $[3/4, 4/3]$. The images were also flipped horizontally with $p=0.5$, and no additional color jitter was performed. For the MobileNet model \citep{howard2017mobilenets}, we used a less aggressive augmentation method, as described in the original paper. In the AlexNet model \citep{krizhevsky2012imagenet}, we used the original augmentation regime.

For all ImageNet models, we followed the training regime by \citet{goyal2017accurate} in which an initial learning rate of $0.1$ is decreased by a factor of $10$ in epochs $30, 60,$ and $80$ for a total of $90$ epochs. We applied a weight decay factor of $10^{-4}$ to every parameter in the network except for those of batch-norm layers.

To fit within our time and compute budget constraints, we used a mild $M=4$ batch augmentation factor for ResNet and MobileNet, and $M=8$ for AlexNet. The ResNet50 model was trained using multiple feed-forwards and gradient accumulations, creating a "Ghost batch normalization" \citep{hoffer2017train} effect, where subsets of 32 images in the batch are normalized separately. We again observe an improvement with all models in their final validation accuracy (Table~\ref{table:val_accuracy}).

The AlexNet model had the most dramatic improvement -- yielding more than $4\%$ improvement in absolute validation accuracy compared to our baseline, and more than $2\%$ than previously best published results \citep{you2017scaling}.

We also highlight the fact that models reached a high validation accuracy quicker. For example, the ResNet50 model, without modification, reached a $75.7\%$ at epoch $35$ -- only $0.6\%$ shy of the final accuracy achieved at epoch $90$ with the baseline model (Figure~\ref{compare_resnet50}). The increase in validation error between epochs $30-60$ suggests that either learning rate or weight-decay values should be altered as discussed by \citet{Zagoruyko2016WRN} who witnessed similar effects. This led us to believe that with careful hyper-parameter tuning of the training regime, we can shorten the number of epochs needed to reach the desired accuracy and even improve it further.

By adapting the training regime to the improved convergence properties of BA, we were able to reduce the number of iterations needed to achieve the required accuracy. Using the same base LR ($0.1$), and reducing by a factor of $0.1$ after epochs 30 and 35 allowed us to reach the same improved accuracy of $76.86\%$ after only 40 epochs. An even faster schedule where the LR is reduced at epochs $15, 20,$ and $22$ yields the previous $75.7\%$ at epoch $23$. 


\subsection{Dropout as intermediate augmentation}
We also tested the ability of batch augmentation to improve results in tasks where no explicit augmentations are performed on input data. An example for this kind of task is language modeling, where the input is fed in a deterministic fashion and noise is introduced in intermediate layers in the form of Dropout \citep{srivastava2014dropout}, DropConnect \citep{dropconnect}, or other forms of regularization \citep{krueger2016zoneout, merity2017regularizing}.

We used the implementation by \citet{merity2017regularizing} and the proposed setting of an LSTM word-level language model over the Penn-Tree-Bank (PTB) dataset. We used a 3-layered LSTM of width 1,150 and embedding size of 400, together with Dropout regularization on both input ($p=0.4$) and hidden state ($p=0.25$), with no fine-tuning.

We used $M=10$, increasing the effective batch-size from 20 to 200. The use of multiple instances of the same samples within the batch caused each instance to be computed with a different random Dropout mask. 

We again observed a positive effect, yet more modest compared to the previous experiments, reaching a $0.2$ improvement in final test perplexity compared to the baseline (see Table \ref{table:val_accuracy}). 

\subsection{Distributed Batch Augmentation}

To support large-scale clusters, we implement distributed BA over TensorFlow and Horovod~\cite{sergeev2018horovod}. The implementation uses decentralized (i.e., without a parameter server) synchronous SGD, and communication is performed using the optimized Message Passing Interface (MPI). We use the maximal number of images per batch per-node, as it provides the best utilization (see Table \ref{tab:tput}).

If we naively replicate a small batch $M$ times on each node, we will degenerate the batch normalization process by normalizing a small set of images with multiple augmentations. Instead, our implementation ensures that every $M$ nodes would load the same batch, so different images are normalized together. Specifically, we achieve this effect by synchronizing the random seeds of the dataset samplers in every $M$ nodes (but not the data augmentation seeds). This also allows the parallel filesystem to detect that the same files are loaded, and broadcast the data after reading it once.

We test our implementation on CSCS Piz Daint, a Cray XC50 supercomputer. Each XC50 compute node contains a 12-core HyperThreading-enabled Intel Xeon E5-2690 CPU with 64 GiB RAM, and one NVIDIA Tesla P100 GPU. The nodes communicate using a Cray Aries interconnect.

\begin{figure}[t]
    \centering
    \includegraphics[width=.48\textwidth,trim={0 0 0 0cm},clip]{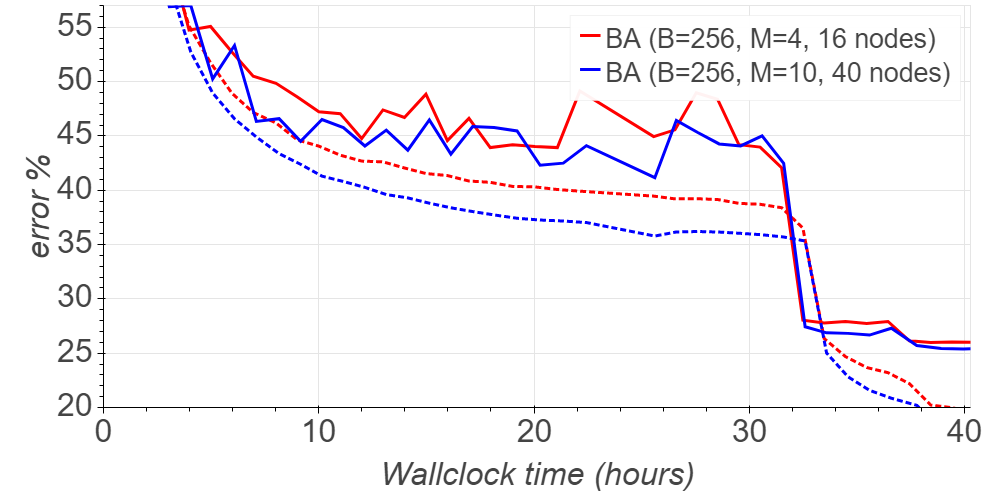}
    \caption{Training (dashed) and validation error over time (in hours) of ResNet50 with $B=256$ and $M=4$ (Red) vs $M=10$ (Blue). Difference in runtime is negligible, while higher batch augmentation reaches lower error. Runtime for Baseline ($M=1$): $1.43\pm 0.13$ steps/second, $M=4$: $1.47\pm 0.13$ steps/second, $M=10$: $1.46\pm 0.14$ steps/second. }\vspace{-1em}\label{compare_distributed}
\end{figure} 

In Figure \ref{compare_distributed}, we plot the training runtime of two experiments on ImageNet with ResNet-50 for $40$ epochs. We test $M=4$ ($16$ nodes) and $M=10$ ($40$ nodes), where each node processes a batch of $64$ images. The plot shows that the difference in runtime for $M=4$ and $M=10$ is negligible, where the larger augmented batch consistently produces increased validation accuracy. The training process uses an augmented batch of $B=256$ distinct samples, with a Ghost Batch Normalization \citep{hoffer2017train} of $32$ images and a standard, but shorter regime (i.e., without adding gradual warmup). This result indicates that BA makes it possible to successfully scale training to an effective batch size of 2,560, without tuning the LR schedule and reduced communication cost due to I/O optimizations. 


\section{Conclusion}
In this work, we have introduced "Batch Augmentation" (BA), a simple yet effective method to improve generalization performance of deep networks by training with large batches composed of multiple transforms of each sample. We have demonstrated significant improvements on various datasets and models, with both faster convergence per epoch, as well as better final validation accuracy.
 
We suggest a theoretical analysis to explain the advantage of BA over traditional large batch methods.
We also show that batch augmentation causes a decrease in gradient variance throughout the training, which is then reflected in the gradient's $\ell_2$ norm used in each optimization step. This may be used in the future to search and adapt more suitable training hyper-parameters, enabling faster convergence and even better performance.

Recent hardware developments allowed the community to use larger batches without increasing the wall clock time either by using data parallelism or by leveraging more advanced hardware. 
However, several papers claimed that working with large batch results in accuracy degradation \citep{masters2018revisiting, golmant2019on}. Here we argue that by using multiple instances of the same sample we can leverage the larger batch capability to increase accuracy. 
These findings give another reason to prefer training settings utilizing significantly larger batches than those advocated in the past.


\bibliography{main}
\bibliographystyle{icml2019}

\newpage
\appendix
\section*{Appendix}

\subsection{Proof of Theorem \ref{thm: maximal eigenvalue}}

We examine the first moment dynamics of  Eq. \ref{eq: SGD-1-1}, by taking its expectation

\begin{equation}
\e\mathbf{w}_{t+1}=\left(\mathbf{I}-\eta\left\langle \mathbf{H}\right\rangle \right)\e\mathbf{w}_{t}\,,\label{eq: First order dynamics-1}
\end{equation}
where 
\[
\left\langle \mathbf{H}\right\rangle \triangleq\frac{1}{N}\sum_{n=1}^{N}\mathbf{H}_{n}
\]

it is easy to see that a necessary and sufficient condition for
convergence of Eq. \ref{eq: First order dynamics-1}
\begin{equation}
\bar{\lambda}_{\max}<\frac{2}{\eta}\:,\label{eq: stability 1-1-1}
\end{equation}
where $\bar{\lambda}_{\max}$ is the maximal eigenvalue of $\left\langle \mathbf{H}\right\rangle $. This is the standard convergence condition for full batch SGD, i.e., gradient descent.

First, to see Eq. \ref{eq: stability 1-2} is a necessary condition
for certain datasets, suppose we have $\mathbf{H}_{n}=0$ in all
samples, except, in a single batch $k$, for which we have
\[
\lambda_{\max}=\max_{\forall\mathbf{v}:\left\Vert \mathbf{v}\right\Vert =1}\mathbf{v}^{\top}\left\langle \mathbf{H}\right\rangle _{k}\mathbf{v}\,,
\]
 In this case, the weights are updated only when we are at batch
$k$. Therefore, ignoring all the batches, the dynamics are equivalent
to full batch gradient descent with the dataset restricted to batch
$k$. Therefore, $\bar{\lambda}_{\max}=\lambda_{\max}$, and we only
have first order dynamics (with no noise). Thus, the necessary and sufficient
condition for stability is Eq. \ref{eq: stability 1-1-1} with $\bar{\lambda}_{\max}=\lambda_{\max}$, which is Eq. \ref{eq: stability 1-2}.

Next, to show Eq. \ref{eq: stability 1-2} is also a sufficient condition
(for all data sets) we examine the second moment dynamics. First we
observe that 
\begin{align*}
\mathbf{w}_{t+1}^{\top}\mathbf{w}_{t+1} & =\mathbf{w}_{t}^{\top}\left(\mathbf{I}-\eta\left\langle \mathbf{H}\right\rangle _{k\left(t\right)}\right)^{\top}\left(\mathbf{I}-\eta\left\langle \mathbf{H}\right\rangle _{k\left(t\right)}\right)\mathbf{w}_{t}\,.\\
 & =\mathbf{w}_{t}^{\top}\left(\mathbf{I}-2\eta\left\langle \mathbf{H}\right\rangle _{k\left(t\right)}+\eta^{2}\left\langle \mathbf{H}\right\rangle _{k\left(t\right)}\left\langle \mathbf{H}\right\rangle _{k\left(t\right)}\right)\mathbf{w}_{t}\,.
\end{align*}
Denoting 
\[
\left\langle \mathbf{H}^{2}\right\rangle \triangleq\frac{1}{N/B}\sum_{k=0}^{N/B}\left\langle \mathbf{H}\right\rangle _{k}\left\langle \mathbf{H}\right\rangle _{k}\,.
\]
Thus, we obtain
\begin{align}
\e\left\Vert \mathbf{w}_{t+1}\right\Vert ^{2} & =\e\left[\mathbf{w}_{t+1}^{\top}\left(\mathbf{I}-2\eta\left\langle \mathbf{H}\right\rangle +\eta^{2}\left\langle \mathbf{H}^{2}\right\rangle \right)\mathbf{w}_{t}\right]\,.\label{eq: second moment dynamics}
\end{align}
Since $\mathbf{H}_{n}$ are all PSDs it is easy to see that if $\mathbf{z}$
is a zero eigenvector of $\left\langle \mathbf{H}\right\rangle $
or $\left\langle \mathbf{H}^{2}\right\rangle $ then it must be a
zero vector eigenvector of other matrix, and also of all $\mathbf{H}_{n}$,
$\forall n$. We denote the null space
\[
\mathcal{V}\triangleq\left\{ \mathbf{v}\in\mathbb{R}^{d}|\left\Vert \mathbf{v}\right\Vert =1,\text{\ensuremath{\left\langle \mathbf{H}\right\rangle }}\mathbf{z}=0\right\} 
\]
and its complement $\bar{\mathcal{\mathcal{V}}}$. From Eq. \ref{eq: second moment dynamics}
a necessary and sufficient condition for convergence of this equation
is 
\begin{equation}
\max_{\mathbf{v}\in\mathcal{\bar{V}}}\mathbf{v}^{\top}\left(\mathbf{I}-2\eta\left\langle \mathbf{H}\right\rangle +\eta^{2}\left\langle \mathbf{H}^{2}\right\rangle \right)\mathbf{v}<1\,.\label{eq: stability 2-1}
\end{equation}
To complete the proof we will show that Eq. \ref{eq: stability 1-2}
also implies Eq. \ref{eq: stability 2-1}, for any $B$.

First we notice that Eq. \ref{eq: lambda max-1} implies that $\forall\mathbf{v}\in\bar{\mathcal{\mathcal{V}}}:$
\begin{align}
\begin{split}
\mathbf{v}^{\top}\left\langle \mathbf{\mathbf{H}}^{2}\right\rangle \mathbf{v} & =\frac{1}{N}\sum_{k=0}^{N/B}\sum_{n\in\mathcal{B}\left(k\right)}\mathbf{v}^{\top}\left\langle \mathbf{H}\right\rangle _{k}\mathbf{H}_{m}\mathbf{v}\\
&\leq\frac{1}{N}\sum_{n=1}^{N}\lambda_{\max}\mathbf{v}^{\top}\mathbf{H}_{n}\mathbf{v}\\
&=\lambda_{\max}\mathbf{v}^{\top}\left\langle \mathbf{H}\right\rangle \mathbf{v}\,.\label{eq: bound 1-2}
\end{split}
\end{align}
Also, since $\lambda_{\max}>\bar{\lambda}_{\max}$, we have 
\begin{equation}
\mathbf{v}^{\top}\left\langle \mathbf{H}\right\rangle ^{2}\mathbf{v}\leq\lambda_{\max}\mathbf{v}^{\top}\left\langle \mathbf{H}\right\rangle \mathbf{v}\,.\label{eq: bound 2-2}
\end{equation}
We combine the above results to prove the Lemma, and $\forall\mathbf{v}\in\bar{\mathcal{\mathcal{V}}}:$
\begin{align*}
& \mathbf{v}^{\top}\left[\left(\mathbf{I}-2\eta\left\langle \mathbf{H}\right\rangle \right)+\eta^{2}\left\langle \mathbf{H}^{2}\right\rangle \right]\mathbf{v}\\
= & 1-2\eta\mathbf{v}^{\top}\left\langle \mathbf{H}\right\rangle \mathbf{v}+\eta^{2}\mathbf{v}^{\top}\left\langle \mathbf{H}^{2}\right\rangle \mathbf{v}\,\\
\overset{\left(1\right)}{\leq} & 1-2\eta\mathbf{v}^{\top}\left\langle \mathbf{H}\right\rangle \mathbf{v}+\eta^{2}\lambda_{\max}\mathbf{v}^{\top}\left\langle \mathbf{H}\right\rangle \mathbf{v}\\
= & 1-\eta\left(2-\eta\lambda_{\max}\right)\mathbf{v}^{\top}\left\langle \mathbf{H}\right\rangle \mathbf{v}\, ,
\end{align*}
where in $\left(1\right)$ we used Eqs. \ref{eq: bound 1-2} and \ref{eq: bound 2-2}.
Given the condition in Eq. \ref{eq: stability 1-2} this is smaller
than $1$, so Eq. \ref{eq: stability 2-1} holds, so this proves the Theorem.

As a side note, we can bound the convergence rate using the last equation.
To see this, we denote $\mathcal{P}_{\bar{\mathcal{\mathcal{V}}}}$
as the projection to $\bar{\mathcal{\mathcal{V}}}$, and
\[
\lambda_{\min}\triangleq\min_{\forall\mathbf{v}\in\bar{\mathcal{\mathcal{V}}}}\mathbf{v}^{\top}\left\langle \mathbf{H}\right\rangle \mathbf{v}
\]
as the smallest non-zero eigenvalue of $\left\langle \mathbf{H}\right\rangle $.
iterating the recursion we obtain that the convergence rate is linear
\begin{equation}
\e\left\Vert \mathcal{P}_{\bar{\mathcal{\mathcal{V}}}}\mathbf{w}_{t}\right\Vert ^{2}\leq\left(1-\eta\left(2-\eta\lambda_{\max}\right)\lambda_{\min}\right)^{t}\e\left\Vert \mathcal{P}_{\bar{\mathcal{\mathcal{V}}}}\mathbf{w}_{0}\right\Vert ^{2}\,.\label{eq: linear convergence-1}
\end{equation}
However, note this bound is not necessarily tight.

\newpage
\subsection{Comparison with longer training}
\begin{table}[ht]
\small
\label{table:compare_ra}
\begin{tabular}{lrrrrr}
\toprule{}\
Network                                       &   Dataset   & M & Baseline  & RA & BA   \\
\midrule
ResNet44                  &   Cifar10 &  10 & 93.07\%  & 93.07\%  & 93.48\% \\
ResNet44 + cutout                             &   Cifar10 &  10 & 93.7\% &  93.8\%  & 94.3\%  \\
WResNet28 + cutout    &  Cifar10  &  10 & 96.6\% & 96.6\%    & 96.95\%  \\
\midrule
AlexNet      &   ImageNet & 8 &  58.25\%  &57.6\%    & 62.31\%   \\
ResNet50                  &   ImageNet &   4 & 76.3\%   & 75.7\%  & 76.86\% \\
\bottomrule
\end{tabular}
\caption{We compare several models over training using BA vs training using larger batches for the same number of iterations (RA). No other training hyper-parameter is modified. in this comparison, both RA and BA ensure the same number of instances seen during training. We show results obtained where the same $M$ multiplier is used for enlarging the epochs and batch size for RA. This again verifies that BA method is useful for generalization beyond only allowing more instances through the course of training.
}
\end{table}

      \begin{figure}[ht]
      \centering
        \includegraphics[width=0.48\textwidth,trim={0 0 0 0cm},clip]{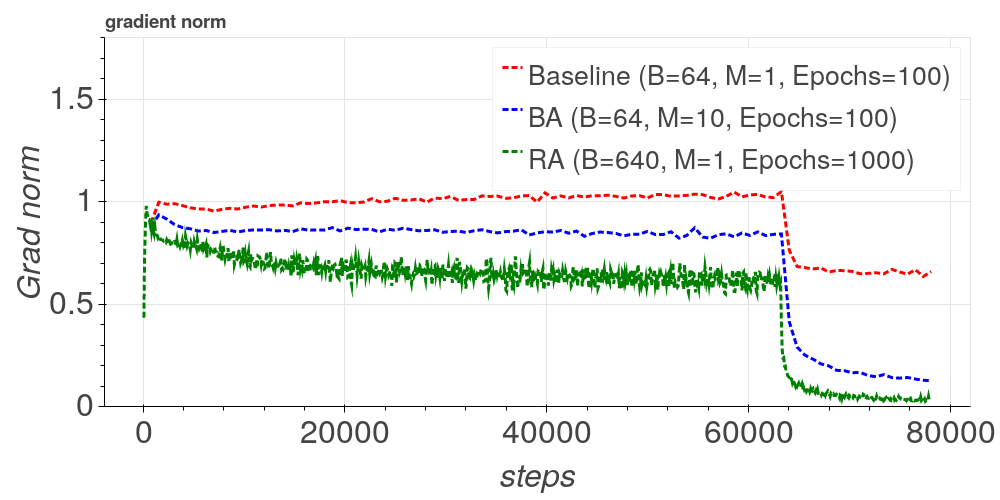}
        \caption{A comparison of gradient norm between 
(1) baseline B=64 training
(2) our batch augmentation (BA) method with M=10
(3) regime adaptation (RA) with B=640 and 10x more epochs. As expected, BA exhibits a gradient norm smaller than Baseline, but larger than large-batch training.}
        \label{ba_ra:grad}
\end{figure}

\end{document}